\documentclass[conference]{IEEEtran}
\IEEEoverridecommandlockouts
\pdfoutput=1
\usepackage{cite}
\usepackage{amsmath,amssymb,amsfonts}
\usepackage{algorithmic}
\usepackage{graphicx}
\usepackage{textcomp}
\usepackage{xcolor}
\usepackage{subcaption}
\usepackage{graphicx}
\begin{document}

\title{Hand Gesture Classification Based on Forearm Ultrasound Video Snippets Using 3D Convolutional Neural Networks\\
\thanks{*Co-first authors}
}

\author{\IEEEauthorblockN{Keshav Bimbraw*}
\IEEEauthorblockA{\textit{Robotics Engineering} \\
\textit{Worcester Polytechnic Institute}\\
Worcester, USA \\
kbimbraw@wpi.edu}
\and
\IEEEauthorblockN{Ankit Talele*}
\IEEEauthorblockA{\textit{Robotics Engineering} \\
\textit{Worcester Polytechnic Institute}\\
Worcester, USA \\
amtalele@wpi.edu}
\and
\IEEEauthorblockN{Haichong K. Zhang}
\IEEEauthorblockA{\textit{Robotics and Biomedical Engineering} \\
\textit{Worcester Polytechnic Institute}\\
Worcester, USA \\
hzhang10@wpi.edu}
}

\maketitle

\begin{abstract}
Ultrasound based hand movement estimation is a crucial area of research with applications in human-machine interaction. Forearm ultrasound offers detailed information about muscle morphology changes during hand movement which can be used to estimate hand gestures. Previous work has focused on analyzing 2-Dimensional (2D) ultrasound image frames using techniques such as convolutional neural networks (CNNs). However, such 2D techniques do not capture temporal features from segments of ultrasound data corresponding to continuous hand movements. This study uses 3D CNN based techniques to capture spatio-temporal patterns within ultrasound video segments for gesture recognition. We compared the performance of a 2D convolution-based network with (2+1)D convolution-based, 3D convolution-based, and our proposed network. Our methodology enhanced the gesture classification accuracy to 98.8 $\pm$ 0.9\%, from 96.5 $\pm$ 2.3\% compared to a network trained with 2D convolution layers. These results demonstrate the advantages of using ultrasound video snippets for improving hand gesture classification performance.
\end{abstract}


\section{Introduction}
Brightness Mode (B-Mode) ultrasound data from the forearm provides a visualization of the physiological mechanisms underlying hand movements and force generation \cite{jacobson2017fundamentals}. This has been used to estimate hand gestures \cite{bimbraw2022prediction}, finger angles \cite{bimbraw2023simultaneous} and finger forces \cite{bimbraw2023estimating}. It has been used for controlling robots \cite{bimbraw2020towards}, prosthetics \cite{hettiarachchi2015new} and virtual reality interfaces \cite{bimbraw2023leveraging}. As ultrasound sensing \cite{frey2022wulpus} and processing \cite{bimbraw2024forearm} becomes smaller and smaller, there is a need to further improve the performance of ultrasound-based hand gesture classification. Most prior research has focused on processing 2-Dimensional (2D) B-mode ultrasound data for this purpose \cite{bimbraw2020towards, bimbraw2023simultaneous, mcintosh2017echoflex}. Notably, convolutional neural networks (CNNs) have been used for forearm ultrasound based gesture classification \cite{bimbraw2022prediction, bimbraw2023simultaneous}. These networks extract spatial features from the ultrasound images during training to optimize their parameters. When the gestures are acquired dynamically, as in \cite{bimbraw2022prediction, bimbraw2023simultaneous}, processing data in a 2D fashion doesn't leverage the advantages of the hand movement undertaken over time. 

Spatiotemporal convolutions (referred to as (2+1)D convolutions) have been used to design neural networks for spatial and temporal feature based action classification \cite{tran2018closer}. Such convolutions have been used to design neural networks used for classification and segmentation tasks. Rehman et al. used 3D CNN for brain tumor detection and classification \cite{rehman2021microscopic}. They have also been used for lung cancer screening based on computed tomography (CT) data \cite{yu20202d}. Chen et al. used 3D CNN for segmentation of tumor based on magnetic resonance imaging (MRI) data \cite{chen2018mri}. Ebadi et al. classified lung ultrasound video segments to detect pneumonia \cite{ebadi2021automated}. Rasheed et al. used ultrasound video segments for automated fetal head classification and segmentation \cite{rasheed2021automated}. However, such spatiotemporal techniques have not been used for forearm ultrasound data based gesture classification.These time-varying features can potentially improve gesture detection accuracy. 

This paper proposes a modified (2+1)D convolution neural network model. Its performance is compared with 2D, (2+1)D and 3D convolution based neural network models. By using forearm ultrasound data for 12 gestures acquired from 3 subjects, we show that the proposed approach is superior to 2D, (2+1)D and 3D convolution based networks. Section II describes the data preprocessing and the classifier. Section III describes the experimental design and Section IV describes the results.

\section{Methods}
Forearm ultrasound data from 3 subjects performing 12 hand gestures was used in this study. The subjects alternated between a rest position and the hand gestures. The Vicon motion capture system was used to acquire ground truth finger angle data. Additional information about the data acquisition can be found in \cite{bimbraw2023simultaneous}. 
\subsection{Data Pre-Processing}
The metacarpophalangeal joint angles were calculated from the raw motion capture data for index, middle, ring and pinky fingers. The ultrasound data acquired using a Verasonics system was also preprocessed before training. The joint angles and ultrasound data were used to extract video segments corresponding to each gesture.
\subsubsection{Joint angle calculation}
Motion capture data from the Vicon system tracked the positions of markers placed on the fingers, and this data was used to calculate the angles between the finger joints like in \cite{bimbraw2022prediction}. This raw data was processed and the necessary frames were extracted from the motion capture data. NumPy arrays (.npy files) were created that contained the finger angles for each gesture. This step was crucial for converting raw motion capture data into a format suitable for model training and hand gesture prediction based on finger angles.

\begin{figure}[!h]
    \centering
    \begin{subfigure}{\linewidth}
        \centering
        \includegraphics[width=200pt]{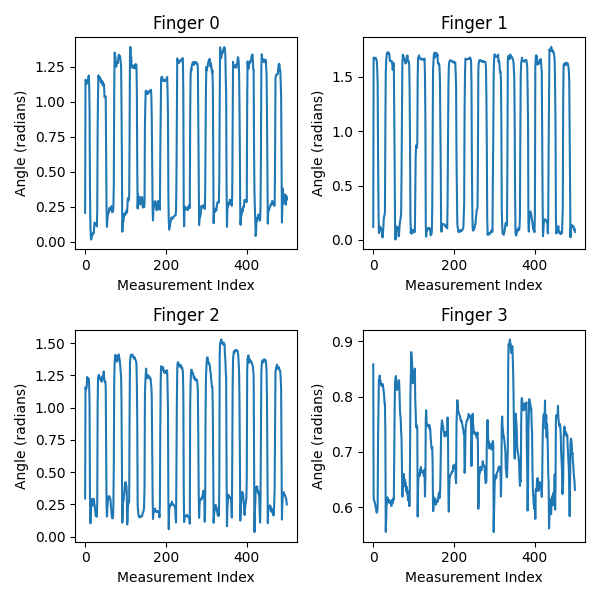}
        \caption{Plot of finger angles. }
        \label{fig:angles}
    \end{subfigure}
    
    \vspace{1em} 
    
    \begin{subfigure}{\linewidth}
        \centering
        \includegraphics[width=150 pt]{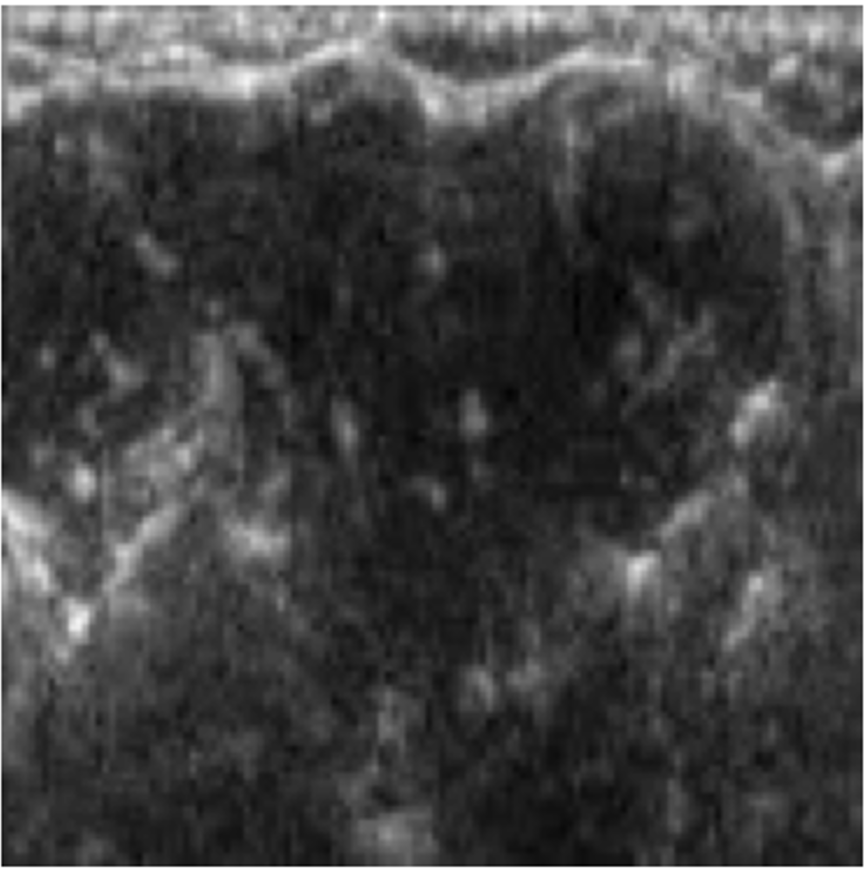}
        \caption{Cropped ultrasound image frame.}
        \label{fig:image}
    \end{subfigure}
    
    \caption{Data visualization: (a) Ground truth for video segment extraction. Plot of finger angles for 500 frames for index (Finger 0), middle (Finger 1), ring (Finger 2), and pinky (Finger 3) fingers, and (b) 224x224 pixel ultrasound image.}
    \label{fig:combined}
\end{figure}

\subsubsection{Ultrasound Images}
For each gesture and subject, 1,400 ultrasound frames were acquired and stored in a single .mat file. The .mat files were converted to .npy arrays and grayscaled, to obtain a final shape of (1400, 636, 256), meaning there were 1,400 frames, each with a resolution of 636x256 pixels per gesture, and subject. The images were cropped to 224x224 pixels to facilitate training by removing redundant information.

\subsubsection{Obtaining video segments}
The video segments were obtained by first calculating the peaks in the finger angle data. This helped estimate the terminal hand position for each gesture. Then, a window of frames surrounding each peak was taken to extract the video segments. This was done for each subject and gesture. 

\begin{figure*}[t]
    \centering
    \includegraphics[width=450pt]{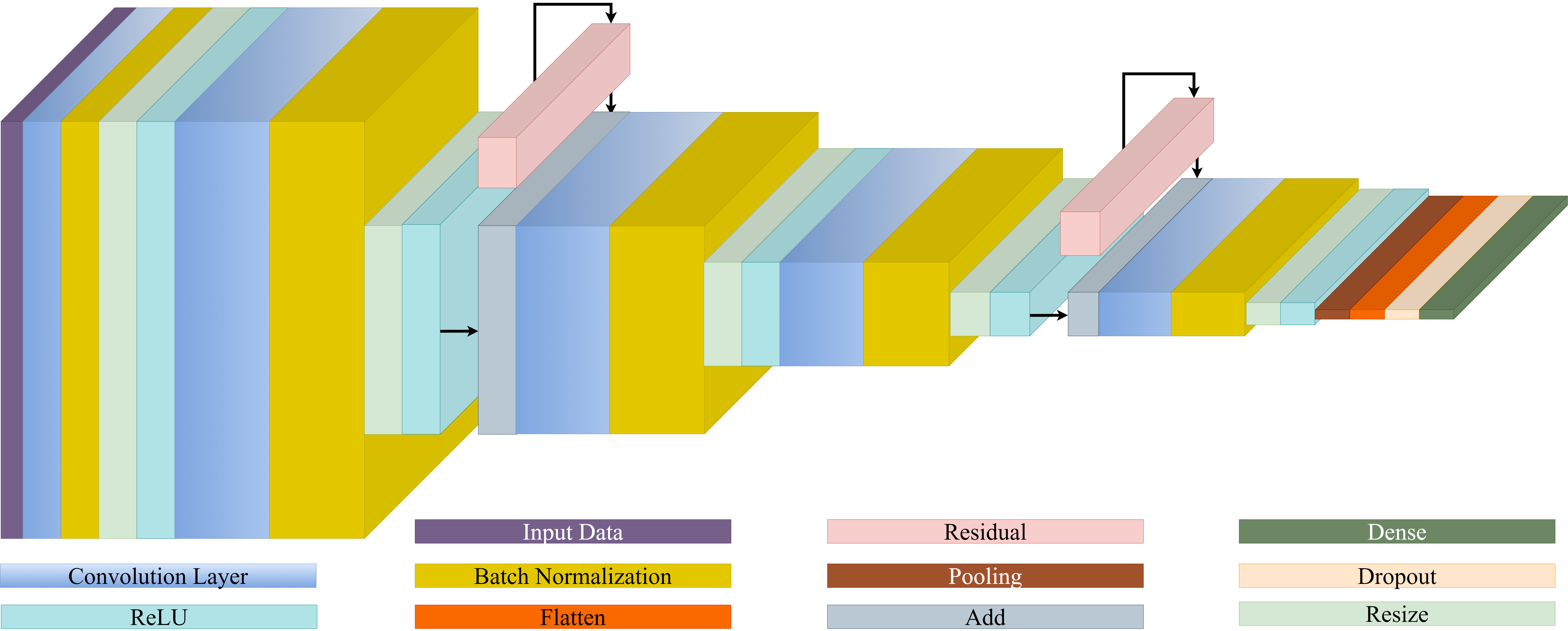}
    \caption{The proposed network with convolution, batch normalization, residual, dense, dropout and pooling layers. Additional operations are indicated.}
    \label{fig:network}
\end{figure*}
\subsection{Classifiers}
2D, 3D and (2+1)D convolutions were used to design neural network based classifiers for this study. These are described as follows.
\subsubsection{2D CNN}
A 2D CNN processes two-dimensional data, such as individual images or image slices. It applies 2D convolutional filters that slide across the height and width of the input, extracting spatial features like edges, textures, and patterns. This architecture is suited for tasks like image classification, object detection, and recognition, where temporal information is irrelevant. However, 2D CNNs cannot capture temporal or depth information, limiting their effectiveness for analyzing sequences or volumetric data.
\subsubsection{3D CNN}
A 3D CNN is designed to handle three-dimensional data, such as video clips or volumetric datasets like MRI scans \cite{chen2018mri}. It uses 3D convolutional filters that slide across height, width, and depth (time or spatial depth), capturing both spatial and temporal features simultaneously. This makes 3D CNNs effective for tasks involving spatiotemporal data, including action recognition, gesture classification, and 3D object detection. However, they are computationally demanding and more prone to overfitting due to the large number of parameters.
\subsubsection{(2+1)D CNN}
The (2+1)D CNN processes 3D data similarly to a 3D CNN but decomposes the process into separate spatial and temporal steps \cite{tran2018closer}. Instead of applying a 3D convolution, it uses a 2D spatial convolution followed by a 1D temporal convolution. This decomposition reduces the number of parameters and improves efficiency. For instance, a 3D convolution with a \(3 \times 3 \times 3\) kernel has significantly more parameters than the (2+1)D version, which uses \(1 \times 3 \times 3\) for spatial convolution and \(3 \times 1 \times 1\) for temporal convolution.

This architecture is particularly useful for tasks that require capturing both spatial and temporal features, such as video-based action recognition or gesture analysis. The reduced computational complexity and enhanced optimization make (2+1)D CNNs more efficient than traditional 3D CNNs. They also allow for better expressiveness by introducing nonlinearities between spatial and temporal convolutions. However, despite reducing parameters, they still require careful tuning and substantial computational resources, especially in deep architectures or large datasets.

\subsubsection{Proposed Network}
The proposed architecture is shown in Figure \ref{fig:network}, and is based on \cite{tran2018closer}. It uses the Conv2Plus1D block, which decomposes 3D convolutions into a 2D spatial convolution followed by a 1D temporal convolution. Initially, the video segment dimensions (depth, height, width) are adjusted using trilinear interpolation. This allows for spatial and temporal resolution throughout the network, balancing computational efficiency with feature preservation. Residual layers consist of pairs of convolution blocks with batch normalization and activation functions. An optional projection is included when input and output dimensions differ, improving gradient flow and stabilizing training in deeper networks.

The network architecture consists of sequential convolution layers with batch normalization, resizing, and ReLU activations \cite{agarap2018deep}. The residual blocks are applied at specific filter sizes, such as 16 and 64 filters, to improve feature learning. The architecture concludes with global average pooling, flattening, and dropout to reduce dimensions and prevent overfitting before the final classification layer. The output layer is a fully connected layer that generates class predictions based on the features extracted by the preceding layers, tailored to the number of target classes. This architecture is designed to capture spatiotemporal features efficiently from video data while maintaining parameter efficiency and robust training dynamics through the use of residual and resizing strategies.

\section{Experimental Design}
The model was trained using data from three subjects, each performing 12 distinct gestures. We used a 20\% test-train split to evaluate the model's ability to generalize across different gestures.

\subsection{Model Training}

Initially, we used a TensorFlow-based video classification model from the original (2+1)D CNN paper \cite{tran2018closer}. However, the TensorFlow data loader was inefficient for our large and complex ultrasound dataset, causing significant performance bottlenecks. To resolve this, we transitioned to PyTorch, enabling the development of a more efficient, customized data loader. This transition improved data throughput and resource utilization, significantly enhancing the training pipeline for large-scale video data.

\subsection{Evaluation}
We compared a slightly modified (2+1)D CNN to the 2D design in \cite{bimbraw2023simultaneous}, a standard 3D model, and the base (2+1)D model in \cite{tran2018closer}. Classification accuracy was used as the primary metric for performance evaluation. Confusion matrices were generated to visualize model performance.

\subsection{Hyperparameters}

Our model's convolution blocks use a (3, 3, 3) kernel size to decompose spatial and temporal dimensions, capturing spatiotemporal features. Padding is set to 'same' to maintain input dimensions. The model uses varying filter sizes, starting from 8 and increasing to 64, to progressively deepen feature extraction. Batch normalization is applied after each convolution block to stabilize learning, and ReLU activations introduce non-linearity.

The model is trained with a batch size of 8, while validation and testing are performed with a batch size of 1. A dropout rate of 0.5 is applied before the final classification layer to reduce overfitting. The training process uses an Adam optimizer with a learning rate of 1e-4 and categorical cross-entropy as the loss function. Data is split into 80\% for training and 20\% for testing, ensuring robust performance evaluation.

\section{Results}

\begin{figure}[h]
    \centering
    \includegraphics[width=200pt]{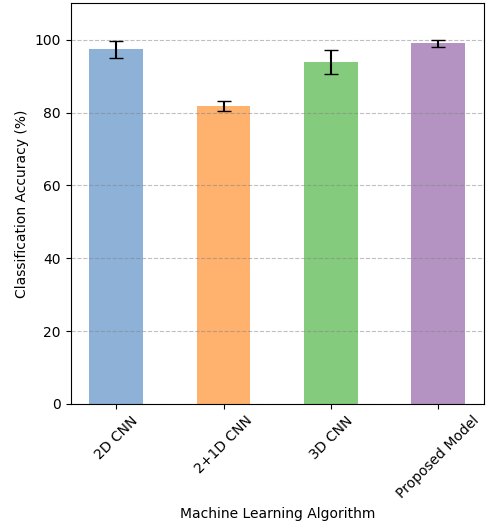}
    \caption{Comparison of classification accuracy across different models. The proposed model achieves the highest accuracy, outperforming both spatial and spatiotemporal baseline architectures.}
    \label{fig:accuracy_comparison}
\end{figure}

We evaluated the proposed model's performance against three baseline architectures: a 2D CNN, (2+1)D CNN, and 3D CNN, using a dataset of 12 hand gestures captured from forearm ultrasound video segments. The classification accuracy for each model is summarized in Figure \ref{fig:accuracy_comparison}. The 2D CNN achieved a classification accuracy of 96.5 ± 2.3\%, showing strong spatial feature extraction but lacking the capacity to capture temporal dynamics. The (2+1)D CNN, which decomposes spatial and temporal convolutions, achieved an accuracy of 86.0 ± 6.1\%. This lower performance likely stems from the model's limited ability to capture the temporal intricacies in ultrasound data. The 3D CNN, which directly models spatiotemporal relationships, outperformed the (2+1)D CNN with a classification accuracy of 92.8 ± 3.1\%, highlighting the significance of temporal feature modeling.

Our proposed model outperformed all baselines, reaching a classification accuracy of 98.8 ± 0.9\%. This superior performance underscores the effectiveness of our spatiotemporal feature extraction approach, combining 2D spatial convolutions with 1D temporal processing while maintaining parameter efficiency. These results emphasize the model's strong generalization across gestures and subjects, showcasing its potential for robust hand gesture classification from ultrasound video data.

\section{Conclusions}
This study demonstrates the effectiveness of spatiotemporal convolution-based neural networks for hand gesture classification using forearm ultrasound video snippets. By incorporating spatiotemporal feature extraction, our proposed model achieved an impressive accuracy of 98.8 ± 0.9\%, significantly outperforming traditional 2D, (2+1)D, and 3D CNN architectures. This advancement highlights the importance of capturing dynamic features in continuous hand movements, suggesting that spatiotemporal approaches can significantly improve gesture classification for human-machine interaction applications. Future work will focus on further optimizing the network architecture and exploring its applicability in real-time gesture recognition systems.

\end{document}